\documentclass{article}

\usepackage{arxiv}

\usepackage[utf8]{inputenc} 
\usepackage[T1]{fontenc}    
\usepackage{hyperref}       
\usepackage{url}            
\usepackage{booktabs}       
\usepackage{amsfonts}       
\usepackage{nicefrac}       
\usepackage{microtype}      
\usepackage{lipsum}
\usepackage{graphicx}
\usepackage{epstopdf}

\title{Namira Soccer 2D Simulation\\ Team Description Paper 2020}

\author{
  Ehsan Asali\\
  Department of Computer Science\\
 University of Georgia\\
  Athens, GA, USA \\
  \texttt{ehsanasali@uga.edu} \\
   \And
 Farzin Negahbani \\
  Department of Computer Science and Engineering\\
  Koç University\\
  Istanbul, Turkey \\
  \texttt{fnegahbani19@ku.edu.tr} \\
   \And
 Shahriyar Bahmaei \\
  Department of Electrical and Computer Engineering\\
  Shiraz University\\
  Shiraz, Fars, Iran \\
  \texttt{bahmaee@cse.shirazu.ac.ir} \\
   \And
  Zahra Abbasi \\
  Department of Electrical and Computer Engineering\\
  Shiraz University\\
  Shiraz, Fars, Iran \\
  \texttt{zahra.abbc97@gmail.com} \\
}

\begin{document}
\maketitle

\begin{abstract}
In this article, we will discuss methods and ideas which are implemented on Namira 2D Soccer Simulation team in the recent year. Numerous scientific and programming activities were done in the process of code development, but we will mention the most outstanding ones in details. A Kalman Filtering method for localization and two helpful software packages will be discussed here. Namira uses agent2d-3.1.1 \cite{akiyama2010agent2d}\cite{akiyama2013helios} as base code and librcsc-4.1.0 as library with some deliberate changes.
\end{abstract}

\keywords{2D Soccer Simulation \and Log Analyzer \and Kalman Filters \and World Model Viewer \and Localization \and Robotics Simulation \and RoboCup}

\section{Introduction}
Soccer 2D Simulation league is one the first robotic leagues in RoboCup Competitions which is a great environment for researchers to invent and apply intelligent algorithms and compete with the best researchers in the field \cite{abreu2012performance}. Numerous teams participate in the WorldCup competition annually which has almost 40 major and junior leagues including simulation and real environments. Moreover, Soccer 2D Simulation league has participants from varied countries and universities. From the most famous teams we can mention Helios \cite{helios2018}, Cyrus \cite{zarecyrus2017}\cite{zarecyrus2019}, Gliders \cite{gliders2019}, FRA-UNIted \cite{fra2020}, Namira \cite{asali2018namira}, and Razi \cite{noohpishehrazi} that have multiple titles in different RoboCup competitions.\\
Namira 2D Soccer Simulation team consists of current and previous students of Shiraz University and Qazvin Islamic Azad University (QIAU). Some of the members were previously working as a team in Shiraz \cite{asali2016shiraz} and Persian Gulf 2D Soccer Simulation Teams \cite{asali2017persian} in World Cup 2016 and 2017 and some recently added students who study Software \& Hardware Engineering at Shiraz University and QIAU. Totally, Namira's members achieved $1^{st}$ place in IranOpen 2016 technical challenge, $2^{nd}$ place in ShirazOpen 2018, $5^{th}$ place in IranOpen 2016 and 2017 leagues, $6^{th}$ place in RoboCup WorldCup 2016, $8^{th}$ place in World Cup 2018 \cite{asali2018namira} and a few other achievements \cite{zarecyrus2015}\cite{khayami2014cyrus}\cite{khayami2013m}. The team's research focus is on topics such as Noise Reduction, Opponent Modelling, Behavior \& Strategy Detection \& Selection, software development, Data Mining \& Analysis \cite{asali2016using} and so on.\\
In this paper, we first introduce a noise reduction method for agents' localization based on Autoregression Kalman Filtering. In section three and four we introduce two new software packages which are beneficial for 2D Soccer Simulation community as tools to analyze game data and work on them in a easier way. At last, we talk about future works and paper conclusion in section five.

\section{Kalman Filtering for Localization}

\begin{figure}[ht]
\centering
\includegraphics[width=1.0\textwidth]{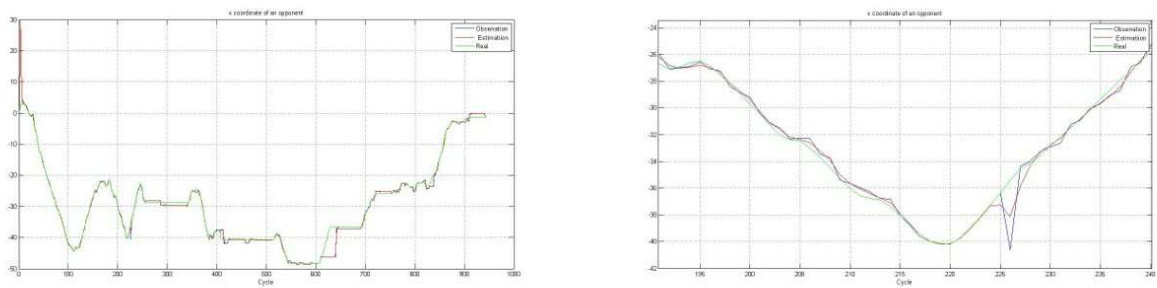}
\caption{Comparison of real, observed and estimated data in 1000 game cycles}
\label{fig:exampleFig1}
\end{figure}

\begin{figure}[ht]
\centering
\includegraphics[width=1.0\textwidth]{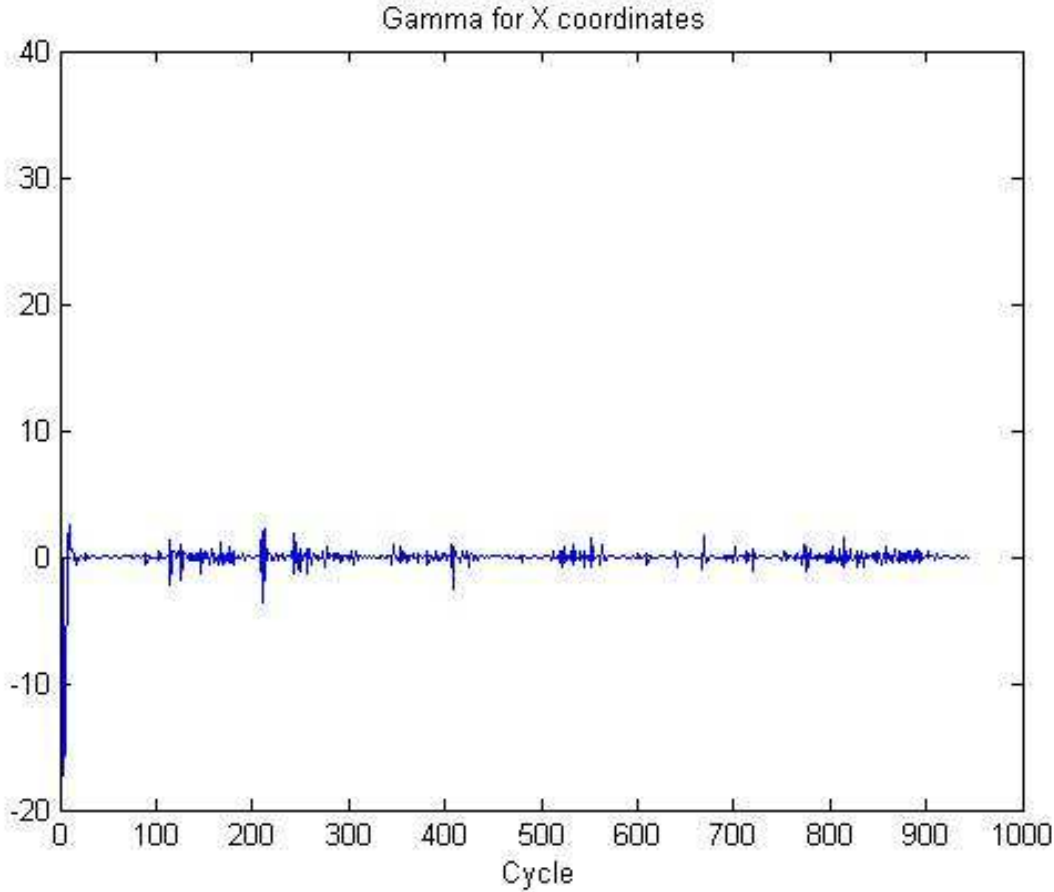}
\caption{Convergence of Gamma parameter over time in Kalman Filter Localization}
\label{fig:exampleFig2}
\end{figure}

Kalman Filtering which is also known as Linear Quadratic Estimation (LQE), is an algorithm that uses a series of measurements observed over time containing statistical noise and other inaccuracies, and produces estimates of unknown variables that tend to be more accurate than those based on a single measurement alone, by estimating a joint probability distribution over the variables for each time frame. Kalman Filtering can be implemented and used in many different ways. In our research, we utilized Autoregressive (AR) Kalman Filter method to fit a model which considers location, velocity, and acceleration to estimate the location of the agents more accurately than what agent already gets from the server. In this model, we assume the movement of the agent has variant velocity and acceleration. The location estimation process works in a way that dedicates more weight to the model than server data in order to handle the effect of acceleration change in agent’s movement during time. As a result, when agent receives noisy location data from server, it compares received data with its own model and tunes the data in a way that huge noises would be alleviated and agent’s belief of the world state will not change dramatically in a short amount of time. \\
As it is shown in the left image of Fig.~\ref{fig:exampleFig1}, we compared x coordinate of an agent in the first 1000 cycles of a match while observation, estimation and real data are depicted in diagram. To see the difference on a closer scrutiny, the right image in Fig.~\ref{fig:exampleFig1} declares that estimated x coordinate is almost always closer to the real data in comparison with observation data which comes from server. The model smooths the dramatic changes in coordinates and does not let the agent to be confused too much about its own or other agents’ locations.
To check the error of our method, we have used Gamma parameter of our Kalman Filter and the results are drawn as a diagram in Fig.~\ref{fig:exampleFig2}.

\section{Log Analyzer}

The Process of analyzing soccer matches is completely vital to find weaknesses and strengths, however, such process is really time consuming without special tools. Traditionally, the most common way to check the performance of the team, specifically after a change in the code, was to observe the game logs by programmer or other experts. In some cases, the changes in the code are very subtle and we need more than a few matches to see the difference. In such cases, we need a great amount of time to play multiple matches and analyze them one by one. Furthermore, altering one behavior may affect other behaviors which may not be detectable by the expert at first glance. Consequently, we need a tool to help us holding matches against different rivals, analyze both teams’ behaviors and give us a feedback so that we can decide if our changes had a positive or negative effect in comparison to our previous code versions.\\
In the previous year, our team proposed a Tournament Planning and Analyzing Software that had a limited built-in version of Log Analyzer that could be used just with our TPAS software. Therefore, we decided not only to improve Log Analyzer, but also to release a standalone version with new features. Both Log Analyzer and TPAS source codes and instructions are provided on github. Since Log Analyzer is cross platform, we can utilize it in all operating systems because it only uses game logs as input data.\\
Log Analyzer has a built-in parser which is capable of parsing rcl and rcg log files and it can both return python objects and text files which contain match facts and other information that can be extracted from the game. The resultant information includes, but not limited to, the number of, shoots, passes, tackles, catches (by goalkeepers), turns, kicks, and so on. In some cases, like shoots or passes, it can also differ correct actions from wrong ones and it calculates the accuracy of the action which is quite useful. Furthermore, we can define a certain region in the field and get number of each event there which gives developers a free hand to focus on special regions. It can also calculate all parameters individually for each player that helps to know whether a target player can do well enough and to detect which players play crucial roles in the game. The accuracy of action detection is almost 99 percent for most actions like shoot and pass. Another important data is possession which indicates percentage of ball possession for each team and player in different regions of the field. This will greatly help experts to balance between their defense, midfield and offense regions and countless other information and knowledge can be gained by having such data. It should be mentioned that we can use Log Analyzer both as a python module or as a script to get reports which again can be both as a python object or a text file. 

\section{World Model Viewer (WMV)}

\begin{figure}[ht]
\centering
\includegraphics[width=1.0\textwidth]{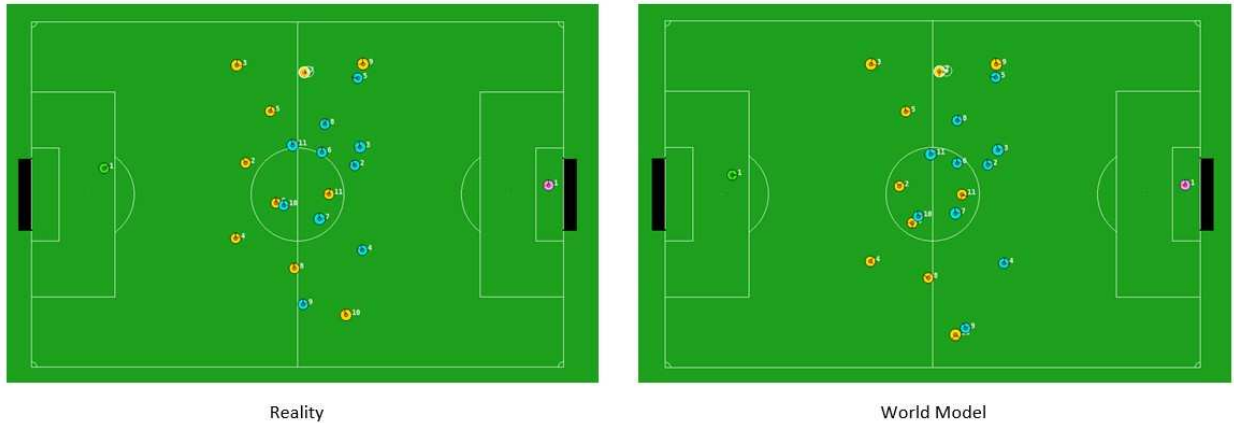}
\caption{Difference between real and world model data for an agent}
\label{fig:exampleFig4}
\end{figure}

The 2D Soccer Simulation environment is partially observable and agents receive noisy data from their observation. To check the difference between real and observed data, typically, we have to print out agent's data and read text outputs. The faster and more convenient way is to check visualized data of agents' observation which was not available through previous tools in the community. 2D Soccer Simulation teams usually use Logplayer software to watch game log files which what they see is based on the real data not what agents can have access to. In order to see agents world model without reinventing the wheel, we developed a software package that uses agents' world model data to generate a new log file for each player and then, we can use Logplayer to see each player's belief state in each cycle of the game. \\
World Model Viewer (WMV) uses Namira Log Analyzer python module, Logplayer and some other python scripts to provide desired data which is players' world model. We embedded some code in agent's code to output its world model in a dedicated file and run a python script to convert previous output data into a readable file for Logplayer. Then, we can watch the game from the eyes of each agent by running its data file using Logplayer. \\
This method has its own pros and cons. We can use Logplayer special features to see object details such as agents' velocity, position, size, type, etc. in a visual manner which is preferable most of the time. Nevertheless, we generate log file for each agent which is 11 files in total; and we can only use this tool when the game is finished. We already replace real data when agent does not have data for a cycle and we are working on that to make it more convenient for users. Figure~\ref{fig:exampleFig4} depicts agent's world compared to real data in an arbitrary cycle of the match.

\section{Conclusion and Future Work}
This paper was a brief description of our scientific and technical efforts as Namira 2D Soccer Simulation team. we mentioned localization enhancement using Kalman Filtering and two novel software, LogAnalyzer and World Model Viewer (WMV), which can be accessed through github freely. We are going to integrated our new software with our past introduced software in previous competitions and publish them as a package in the near future. We also need to make improvement in our WMV software to have online view over agent's world model in Logplayer. Implementing learning algorithms for behavior modelling and detection is another goal of the team in for the future competitions.

\bibliographystyle{unsrt}  

\bibliography{template}  


\end{document}